\documentclass[lettersize,journal]{IEEEtran}
\usepackage{amsmath,amsfonts}
\usepackage{algorithmic}
\usepackage{array}
\usepackage[caption=false,font=normalsize,labelfont=sf,textfont=sf]{subfig}
\usepackage{textcomp}
\usepackage{stfloats}
\usepackage{url}
\usepackage{verbatim}
\usepackage{graphicx}
\usepackage{booktabs}
\usepackage{color}
\usepackage{soul}
\usepackage[table]{xcolor}% http://ctan.org/pkg/xcolor
\usepackage{tikz}
\usepackage{amssymb}
\usepackage{amssymb}% http://ctan.org/pkg/amssymb
\usepackage{pifont}% http://ctan.org/pkg/pifont
\newcommand{\cmark}{\ding{51}}%
\newcommand{\xmark}{\ding{55}}%
\usepackage{makecell}
\usepackage{xcolor,colortbl}
\newcolumntype{s}{>{\columncolor{yellow}}c}

\def \ie {\emph{i.e.}~}
\def \eg {\emph{e.g.}~}
\def \etc {\emph{etc.}~}
\def \etal {\emph{et al.}~}
\def \vs {\emph{v.s.}~}

\def\BibTeX{{\rm B\kern-.05em{\sc i\kern-.025em b}\kern-.08em
    T\kern-.1667em\lower.7ex\hbox{E}\kern-.125emX}}
\usepackage{balance}

\begin{document}
\title{TreeFormer: a Semi-Supervised Transformer-based Framework for Tree Counting from a Single High Resolution Image}

\author{Hamed Amini Amirkolaee,~%~\IEEEmembership{Staff,~IEEE,}
        Miaojing Shi$^*$,~\IEEEmembership{Senior Member,~IEEE,}
        Mark Mulligan%~\IEEEmembership{Member,~IEEE,}

\thanks{$^*$Corresponding author}
\thanks{Hamed Amini Amirkolaee is with the Department of Informatics, King's College London, London WC2B 4BG, U.K. E-mail: \tt\small hamed.amini-amirkolaee@kcl.ac.uk.}
\thanks{{Miaojing Shi is with the College of Electronic and Information Engineering, Tongji University, Shanghai, 20092, China.  E-mail: \tt\small mshi@tongji.edu.cn.}}% <-this % stops a space
\thanks{Mark Mulligan is with the Department of Geography, King’s College
London, London WC2B 4BG, U.K. E-mail: \tt\small\ mark.mulligan@kcl.ac.uk.}

\markboth{IEEE TRANSACTIONS ON GEOSCIENCE AND REMOTE SENSING}}

\maketitle

\begin{abstract}
Automatic tree density estimation and counting using single aerial and satellite images is a challenging task in photogrammetry and remote sensing, yet has an important role in forest management. In this paper, we propose the first semi-supervised transformer-based framework for tree counting which reduces the expensive tree annotations for remote sensing images. Our method, termed as TreeFormer, first develops a pyramid tree representation module based on transformer blocks to extract multi-scale features during the encoding stage. Contextual attention-based feature fusion and tree density regressor modules are further designed to utilize the robust features from the encoder to estimate tree density maps in the decoder. Moreover, we propose a pyramid learning strategy that includes local tree density consistency and local tree count ranking losses to utilize unlabeled images into the training process. 
Finally, the tree counter token is introduced to regulate the network by computing the global tree counts for both labeled and unlabeled images. Our model was evaluated on two benchmark tree counting datasets, Jiangsu, and Yosemite, as well as a new dataset, KCL-London, created by ourselves. Our TreeFormer outperforms the state of the art semi-supervised methods under the same setting and exceeds the fully-supervised methods using the same number of labeled images. The codes and datasets are available at \emph{\color{magenta}{https://github.com/HAAClassic/TreeFormer}}.
\end{abstract}

\begin{IEEEkeywords}
Tree counting, semi-supervised model, transformer, pyramid learning strategy, remote sensing.
\end{IEEEkeywords}

\section{Introduction}\label{introduction}

\IEEEPARstart{T}{rees} are the pulse of the earth and are vital organisms in maintaining the ecological functioning and health of the planet \cite{weinstein2020neon}. Tree counting using high-resolution images is useful in various fields such as forest inventory \cite{ong2021framework}, urban planning \cite{gomez2010effect}, farm management \cite{caruso2019high}, and crop estimation \cite{shahbazi2014recent}, making it important in photogrammetry, remote sensing, and nature-based solutions to environmental change \cite{mulligan2021environmental}. 

Counting trees using traditional methods such as field surveys based on quadrats is very time-consuming and expensive~\cite{kindermann2008global}. Therefore providing an automatic method in this field can be very helpful and practical \cite{ammar2021deep}. High-resolution aerial and satellite images \cite{kaartinen2008eurosdr, kaartinen2012international,gu2021individual} and light detection and ranging (LiDAR) \cite{ghanbari2021individual,hanssen2021utilizing,liu2013extraction,hao2021hierarchical,wallace2014evaluating} data are the most important sources for tree detection and counting. 3D LiDAR data along with 2D aerial and satellite images can be very effective to achieve  accurate results \cite{weinstein2020neon}. On the other hand, collecting and preparing aerial and satellite images is much less expensive than LiDAR data which makes it worth presenting an automatic method for tree counting using a single high-resolution image \cite{kaartinen2008eurosdr}.

In the last decade, artificial intelligence and especially deep learning have developed greatly and achieved significant success in the field of remote sensing \cite{yao2021tree}. The lack of 3D information in aerial and satellite images makes it difficult to identify and distinguish trees, while the ability of deep neural networks (DNNs) in extracting and  distinguish of the geometric and textural features of trees has made this feasible \cite{ammar2021deep}. Although the supervised learning methods based on DNNs have achieved promising performance in tree counting \cite{weinstein2020neon,yao2021tree,lassalle2022deep,liu2021deep}, a large number of trees must be labeled (\eg in the form of points or bounding boxes) to train these networks, which is very costly and time-consuming, especially for areas where trees are very dense. To solve this problem, a semi-supervised strategy is desirable, in which a limited number of labeled images and a large number of unlabeled images are utilized. Apart from training the model on the labeled data, the main purpose of semi-supervised learning is to design efficient supervision for unlabeled data to include them into the model training \cite{ouali2020semi,tarvainen2017mean,verma2022interpolation}.
The state of the art solutions can be mainly categorized into two classes: pseudo-labeling and consistency regularization. In the first class, the model is trained using the labeled data and is used to generate pseudo labels for unlabeled data. The pseudo labels are then included into the model training for unlabeled data \cite{liu2020semi, sindagi2020learning}. 
In the second class, the model is trained on both labeled and unlabeled data using a supervised loss and a consistency loss, respectively. 
The supervised loss is task-related while the consistency loss is normally applied as a regulator to force the agreement between results obtained from differently-augmented unlabeled images \cite{tarvainen2017mean,ouali2020semi}. In semi-supervised object counting \cite{gao2022s,liu2018leveraging,liu2019exploiting}, a ranking constraint is often employed to investigate the count relations between the super- and sub-regions of an image. 

In this paper, we for the first time propose a semi-supervised framework for tree counting, namely TreeFormer. It is built upon a transformer structure. 
In recent years, the transformer has attracted a lot of attention in our community and has had very promising results in many visual tasks \cite{dai2021up,dosovitskiy2020image}. This is due to their strong capacity to aggregate local information using self-attention and propagate representations from lower to higher layers in the network.
We base our network encoder on a pyramid vision transformer (PVT) \cite{wang2021pyramid} to extract robust multi-scale features. A contextual attention-based feature fusion module is introduced to utilize these features in the network decoder. We develop the decoder to produce pyramid predictions by adding a tree density regressor module after each scale feature. In addition, we notice the CLASS token in the PVT gathers global information from all patches for image classification \cite{dosovitskiy2020image}. 
Inspired by it, we design a new tree counter token to estimate the global tree count at each scale of our network encoder. 

Our network optimization follows a pyramid learning strategy, \ie pixel-level, region-level, and image-level learning. {For labeled data, the estimated tree density maps are compared with ground truth using \emph{pixel-level} distribution matching loss.} To effectively leverage the unlabeled data, we introduce two \emph{region-level} losses:  local tree density consistency loss and local tree count ranking loss. 
The local tree density consistency loss is proposed to encourage the tree density predictions from the same local region over different scales to be consistent for a given input. In order to encourage the invariance of the model’s predictions, different scales are perturbed with noises.
The local tree count ranking loss is proposed to control the tree numbers in different local regions of the tree density map so that a super-region contains equal or more trees than its sub-region in an image. 
Finally, the network is optimized on the \emph{image-level} multi-scale tree counts predicted by the tree counter tokens. For a labeled image, these predictions are directly compared with the ground truth tree count. For an unlabeled image, we average these predictions to serve as a global pseudo supervision, which encourages the multi-scale outputs to be close for the same image.
In summary, the main contribution of this work, TreeFormer, is threefold:

\begin{itemize}

\item For network architecture, a \textit{pyramid tree feature representation} module is employed for the encoder and a \textit{contextual attention-based feature fusion} module is designed to utilize the pyramid features for the decoder. A \emph{tree density regressor module} and \textit{tree counter token} are introduced to predict the tree density map and global tree count at each scale, respectively.

\item For network optimization, a \emph{pyramid learning strategy} is designed.  Specifically, a scheme of learning from unlabeled images using \textit{local tree density consistency} and \textit{local tree count ranking} losses on the region-level is emphasized; an image-level \emph{global tree count regularization} based on the global predictions from tree counter tokens is also highlighted.

\item For network benchmarking, we create a  new tree counting dataset, \emph{KCL-London}, from London, UK. 

This dataset contains 921 high-resolution images that are gathered by manually digitizing Google Earth imagery. Individual tree locations in the images are manually annotated.

\end{itemize}

{We conduct extensive experiments on three datasets, \ie Jiangsu \cite{yao2021tree, liu2021deep}, Yosemite \cite{chen2022transformer} and KCL-London. Our method outperforms the state of the art significantly.}

\section{Related Works}

We survey the related works in two subsections:  object counting and tree counting.

\subsection{Object counting}\label{ObjectCounting}

Object counting methods have been used in various fields such as human crowds, car \cite{amato2019counting,biswas2017automatic}, cells \cite{falk2019u,xie2018microscopy} and, trees \cite{weinstein2020neon,yao2021tree}. The challenges of object counting include scale variation, severe occlusions, appearance variations, illumination conditions, and perspective distortions
\cite{chattopadhyay2017counting,stewart2016end,du2022redesigning,jiang2022multi}. Many methods proposed in the field of object counting are related to crowd counting~\cite{liu2018leveraging,gao2022s,meng2021spatial,zhao2020active,sam2020locate,wang2022hybrid,liang2022transcrowd}. Below we discuss these methods in two parts including fully supervised and partially supervised methods.

\subsubsection{Fully supervised methods}\label{FullySupervisedMethods} 
These methods usually convert the point-level annotations of object centers into density maps using Gaussian kernels and utilize them as ground truth.
They achieve good performance via training with a large amount of annotated data.  In order to solve the challenge of scale variation in crowd counting, multi-column/-scale networks are popular architectures to choose \cite{boominathan2016crowdnet,yang2020embedding,zhang2016single}. {The visual attention mechanism is also effective to address the problem of scale variation and background noise in crowded scenes \cite{jiang2020attention,wang2022hybrid}.} In addition, employing auxiliary tasks such as localization \cite{khan2021scale,li2021approaches,sam2020locate}, classification \cite{shi2019counting,sindagi2019ha}, and segmentation \cite{gao2019pcc,zhao2019leveraging} are useful to improve the counting performance. 

\subsubsection{Partially supervised methods}\label{PartiallySupervisedMethods} 
Recently, researchers tried to reduce the need for labeled training data by developing weakly/semi-supervised methods. Semi-supervised methods alleviate the annotation burden by using additional unlabeled data that can help achieve high accuracy with a smaller number of labeled data only. For instance, Liu ~\etal \cite{liu2018leveraging} introduced a pairwise ranking loss to estimate a density map using a large number of unlabeled images. Wang \etal \cite{wang2020nwpu} reduced the need for annotation by combining real and synthetic images. Another strategy is to estimate pseudo labels of unlabeled images and use them in a supervised network to improve the accuracy of the results \cite{sindagi2020learning,meng2021spatial}. Recently, Zhao \etal \cite{zhao2020active} proposed an active labeling strategy to annotate the most informative images in the dataset and learn the counting model upon both labeled and unlabeled images \cite{zhao2020active}. Sam \etal \cite{sam2019almost} presented a stacked convolution autoencoder based on the grid winner-take-all paradigm in which most of the parameters can be learned with unlabeled data.

Weakly-supervised methods aim to use global counts instead of point-level annotations for model learning~\cite{lei2021towards, yang2020weakly,liang2022transcrowd}. For example, Yang \etal \cite{yang2020weakly} presented a weakly-supervised counting network, which directly regresses the crowd numbers without location supervision. They utilized a soft-label sorting network along with a counting network to sort  images according to their crowd numbers. 

\begin{figure*}[!t]
\centering
\includegraphics[width=7in]{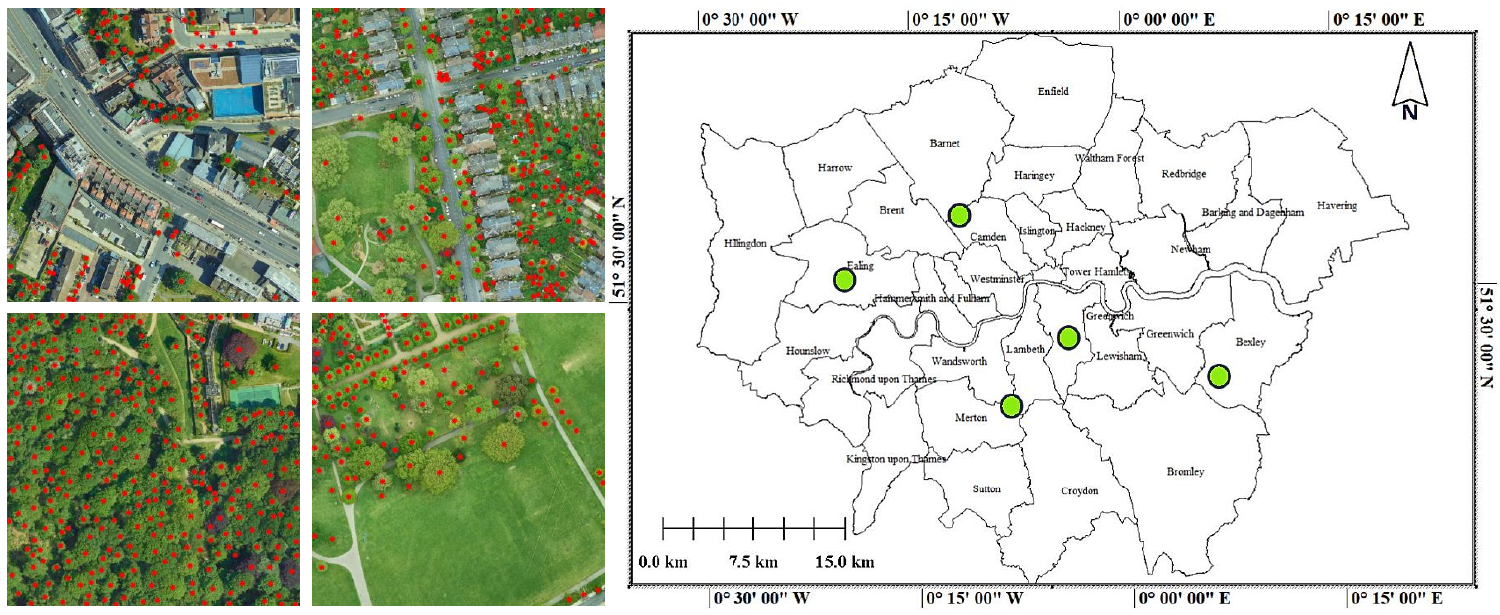}
\caption{Some samples of the annotated  images (Left: red dots) and their locations in London (Right: green circles).} 

\label{fig:dataset}
\end{figure*}

\subsection{Tree counting}\label{TreeCounting} 
Counting trees in the dense tree canopy where trees are very close and sometimes interlocking becomes much more difficult than counting other objects such as humans, cars, cells, \etc In other words, trees can be in continuous form from the top view and their separation using a single image is very complex.

Traditionally, the area where trees exist is detected, then algorithms such as region growing \cite{culvenor2002tida}, watershed segmentation \cite{wang2010multi}, and template matching \cite{wang2019automatic} are used to segment and count trees. In these methods, suitable features are selected and produced by analyzing the spectral, textural, and geometrical characteristics of trees. The accuracy of these methods is dependent on the strength of the handcrafted features that were manually engineered by researchers. 
In regions with dense and complex tree cover, their accuracies are not satisfactory.

Recently, the successful performance of deep neural networks (DNNs) in object detection~\cite{yi2021probabilistic,wu2020recent,diwan2022object} has inspired researchers to adapt these algorithms for the detection and counting of trees. In these networks, suitable features are automatically learned by the network. The widely used DNNs for tree counting are either based on detection~\cite{weinstein2020neon,ammar2021deep,machefer2020mask,weinstein2019individual,zheng2020cross} or density estimation  \cite{yao2021tree,liu2021deep,chen2022transformer,osco2020convolutional}. 
\subsubsection{Detection-based methods} These methods count the number of trees in each image by identifying and localizing individual trees with bounding boxes. Machefer \etal~\cite{machefer2020mask} utilized a Mask R-CNN for tree counting from unmanned aerial vehicle (UAV) images. They focused on low-density crops, potatoes, and lettuce, and employed a transfer learning technique to reduce the requirement for training data. Zheng~\etal~\cite{zheng2020cross} presented a domain adaptive network to detect and count oil palm trees. They employed a multi-level attention mechanism including entropy-level attention and feature-level attention to enhance the transferability of different domains. Weinsteinl~\etal~\cite{weinstein2020neon,weinstein2019individual} produced an open-source dataset for tree crown estimation at different sites across the United States. They show that deep learning models can leverage existing LiDAR-based unsupervised delineation to generate training data for a tree detection network \cite{weinstein2019individual}. Ammar ~\etal~\cite{ammar2021deep} compared the performance of different networks such as Faster R-CNN, YOLOv3, YOLOv4, and EfficientNet for the automated counting and geolocation of palm trees from aerial images. Lassalle \etal \cite{lassalle2022deep} combined a DNN with watershed segmentation to delineate individual tree crowns.

\subsubsection{Density estimation based methods}\label{DensityEstimationBasedMethods}
The performance of the detection-based methods is unsatisfactory when encountering the situation of occlusion and background clutter in extremely dense tree regions. The density estimation-based methods learn the mapping from an image to its tree number which avoids the dependence on the detector and often has higher performance. 
A density map is normally produced by convolving a Gaussian function with specified neighborhood size and sigma at every annotated tree location in an image. The integral of the density map is equal to the number of trees in the image. Chen and Shang \cite{chen2022transformer} combined a convolutional neural network (CNN) and transformer blocks to estimate the density map. 
Osco \etal \cite{osco2020convolutional} employed a DNN to estimate the number of citrus trees by predicting density map from UAV multispectral imagery. They also analyzed the effect of using near infrared band on the achieved results. 
Yao \etal \cite{yao2021tree} constructed a tree counting dataset using four GF-II images and utilized a two-column DNN based on VGGnet and Alexnet for tree density estimation. Liu \etal \cite{liu2021deep} proposed a pyramid encoding–decoding network, which integrates the features from the multiple decoding paths and adapts the characteristics of trees at different scales.

\medskip
In general, there is not much research on tree density estimation, even though they mainly use common and basic networks in deep learning \cite{liu2021deep,osco2020convolutional,yao2021tree}. Also, the existing algorithms in this field are supervised methods, while it is vital to provide a semi-supervised method with an efficient structure due to the lack of annotated training data in this field.

\begin{figure*}[!t]
\centering
\includegraphics[width=7in]{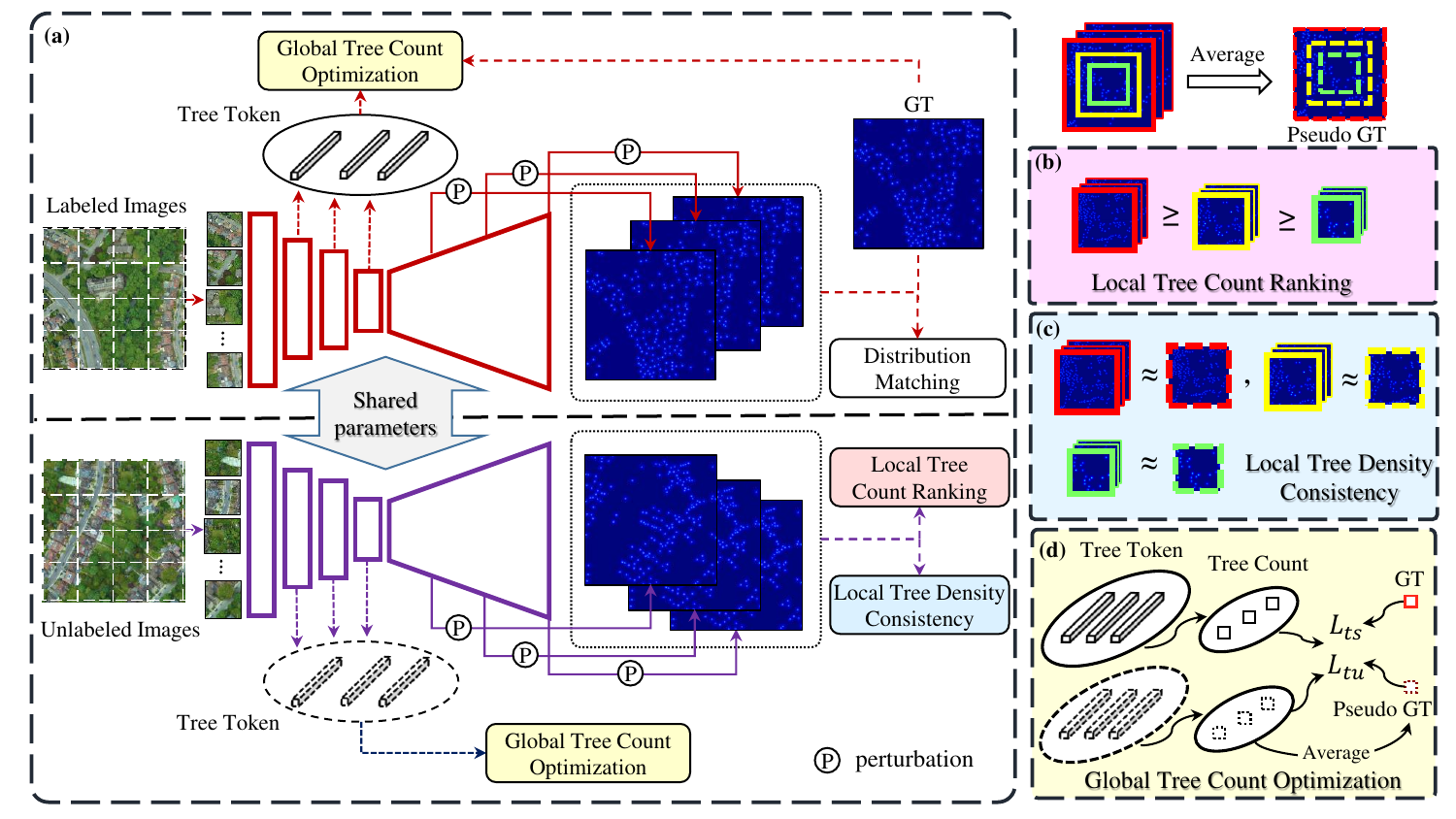}
\caption{Overview of the proposed TreeFormer framework. (a) In the top row, the estimated results of labeled images in different scales are optimized with ground truth (GT) using the distribution matching loss. In the bottom row,  the estimated results of unlabeled images in different scales are optimized with the local tree density consistency and local tree count ranking losses. Moreover, the tree counter tokens are used to predict global tree numbers of images and compare them to either GT for labeled images or mean prediction for unlabeled images. (b) The structure of the local tree count ranking loss. (c) The structure of the local tree density consistency loss. (d) The structure of the global tree count optimization. }
\label{fig:overview}
\end{figure*}

\section{Data source}\label{DataSource}
\subsection{Area}\label{StudyArea}
The area is focused on London, the United Kingdom 
Collating data about London's urban forest is challenging due to the number of landowners and managers involved.
This city contains trees with different types, sizes, shapes, and densities which are challenging to detect and count using traditional remote sensing approaches. Some trees are isolated on streets and others are together in small recreational areas or large areas of ancient forested parkland. Backgrounds are sometimes pavement and sometimes grassland, water, or other trees. London also has different tree species such as Apple, Ash, Cherry, Hawthorn, Hornbeam, Lime, Maple, Oak, Pear, \etc which have different canopy shapes and characteristics. 
In addition to the above varieties of trees, there are also trees with different arrangements in the areas of the city. For example, in central areas of the city, trees have a low density and are located at a greater distance from each other; while the density of trees is very high at the edge of the city.

\subsection{Labels}\label{Labels}
The required high-resolution images are gathered and stitched together from Google Maps at 0.2 m ground sampling distance (GSD). The gathered images are divided into images with 1024 × 1024 pixels. 
To aid the identification of tree locations and numbers of selected images, we employed the accessible tree locations of London in London Datastore website\footnote{https://data.london.gov.uk/dataset/local-authority-maintained-trees}. 
Although these data show the locations and species information for over 880,000 of London's trees, the data mainly contains information on trees in the main streets and does not cover trees that are dense between houses or parks. We manually annotated the latter.
To this end, Global Mapper as geographic information system software is used to annotate the center of each tree. The tree labels are rasterized and converted to JPG format with a resolution compatible with the image data.

\subsection{Characteristics}\label{Characteristics}
The prepared dataset, termed as KCL-London,  consists of 613 labeled and 308 unlabeled images. 95,067 trees were annotated in total in the labeled images.
The tree number in these images varies from about 4 in areas with sparse covers to 332 in areas with dense covers. These images are gathered from different locations that represent a range of different areas across London. In Fig. \ref{fig:dataset} the selected locations of prepared images with annotations are presented.

\section{Methodology}\label{sec:Methodology}
\subsection{Overview}\label{sec:OverviewS}

In this paper, a semi-supervised framework is proposed to estimate the density map of trees from a remote sensing image. An overview of the designed framework is presented in Fig. \ref{fig:overview}. Our network has an encoder-decoder architecture based on transformer blocks. A pyramid tree feature representation (PTFR) module is developed in the encoder to extract multi-phase features from the input image (Sec. \ref{sec:PyramidTreeFeatureRepresentation}). A contextual attention-based feature fusion (CAFF) module is introduced  to utilize the pyramid features in the decoder (Sec. \ref{sec:ContextualAttentionFeatureFusion}). 
Afterwards, the tree density map is estimated in each scale of the decoder using the designed tree density regressor (TDR) module  (Sec. \ref{sec:TreeDensityRegressor}). Besides, a tree counter token (TCT) is proposed to compute the number of trees in each phase of the encoder (Sec. \ref{sec:TreeCountingToken}). 
 
For the labeled data, a supervised distribution matching loss is employed to train the network (Sec. \ref{sec:PixelLevelLearning}).
The same architecture with shared parameters is used for unlabeled data, while the proposed local tree density consistency and local tree count ranking losses are utilized to assist the network to achieve more accurate results (Sec. \ref{sec:RegionLevelLearning}). A global tree count regularization that optimizes the global tree count predictions from the tree counter tokens is applied to both labeled and unlabeled data (Sec. \ref{sec:ImageLevelLearning}). The loss functions used for labeled and unlabeled data are applied to the pyramid estimations of the proposed model.

\subsection{TreeFormer framework}\label{sec:TreeFormerFramework}
\begin{figure*}[!t]
\centering
\includegraphics[width=7in]{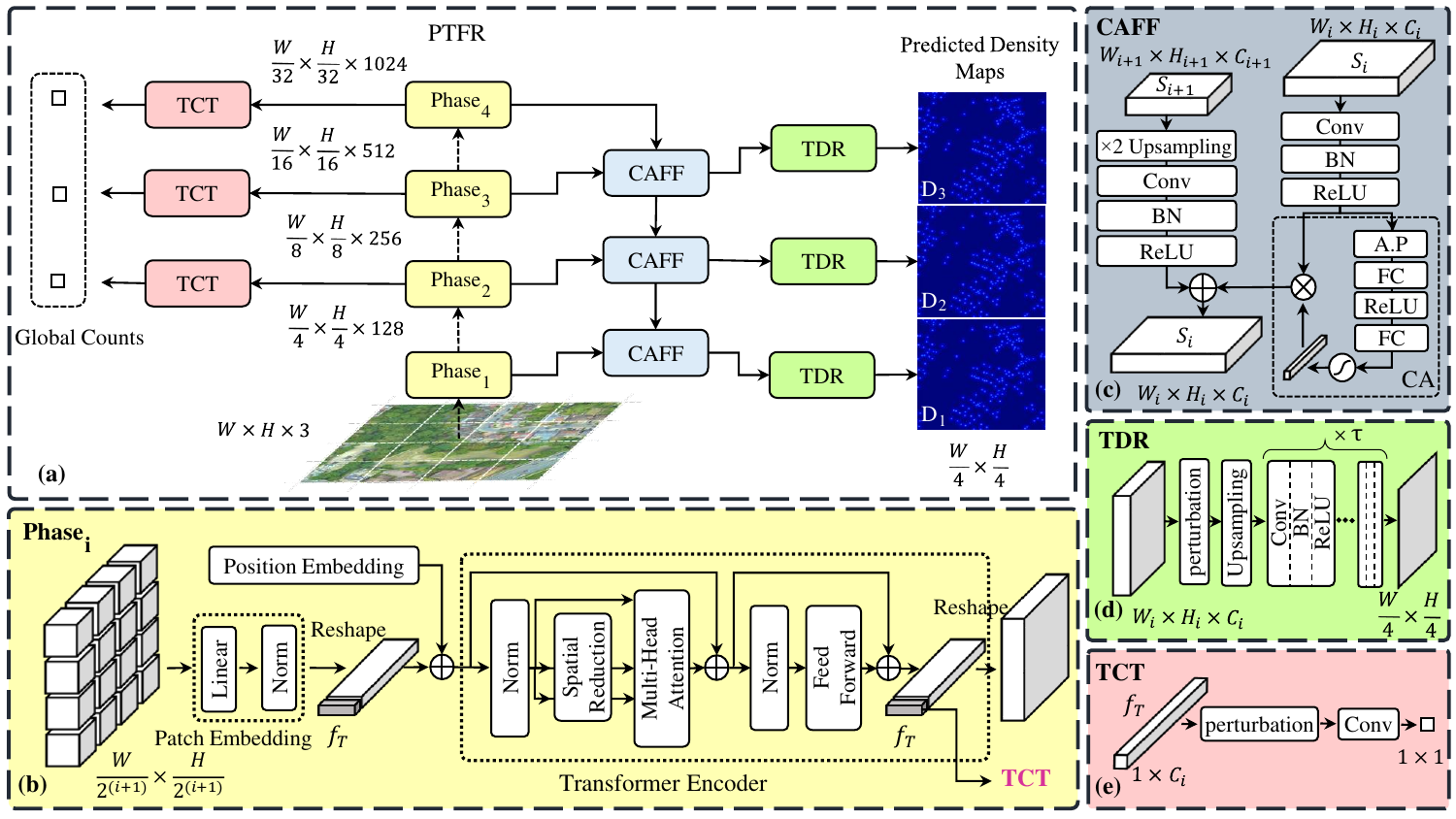}
\caption{(a) The details of the encoder-decoder architecture in TreeFormer. Given the input image, multi-phase features are first extracted through the PTFR module in the encoder. The CAFF and TDR modules in the decoder fuse feature maps over scales to acquire more robust feature representations and estimate the tree density map in each scale, respectively. (b) {A certain phase of PTFR extracts the feature map based on the transformer encoder}. (c) The CAFF module fuses the coarser-resolution feature of the decoder with the finer-resolution feature of the encoder. (d) The TDR module estimates the tree density map in each scale. (e) The TCT module estimates the global number of trees from each scale of the encoder and uses it for global optimization.} 
\label{fig:modules}
\end{figure*}

In this section, we introduce the pyramid tree feature representations and the tree counter tokens for the encoder of our TreeFormer; the contextual attention-based feature fusion modules, and the tree density regressor modules for the decoder of our module. They are also illustrated in Fig.~\ref{fig:modules} in details.

\subsubsection{Pyramid Tree Feature Representation}\label{sec:PyramidTreeFeatureRepresentation}
We develop the PTFR based on the pyramid vision transformer (PVT) \cite{wang2021pyramid} to effectively extract multi-phase features in the encoding process. 
The PVT divides the image into $4\times4$ non-overlapping patches as input. The PTFR is obtained by applying convolutional layers with different strides in each phase of the PVT. Suppose $W$ and $H$ represent the width and height of the input image, a set of feature maps including phase 1: $\frac{W}{4}\times\frac{H}{4}\times128$, phase 2: $\frac{W}{8}\times\frac{H}{8}\times256$, phase 3: $\frac{W}{16}\times\frac{H}{16}\times512$, phase 4: $\frac{W}{32}\times\frac{H}{32}\times1024$, are generated in PTFR.
The achieved feature map in each phase is both fed to the CAFF module, specified next, and utilized as input (half-sized) for the next phase (Fig.\ref{fig:modules}a). Notice following \mbox{\cite{wang2021pyramid}} we half the resolution of the feature map while double the number of channels at each scale.

In the $i$-th phase, as illustrated in Fig.~\ref{fig:modules}b, the input image is divided into $\frac{W}{2^{i+1}}\times\frac{H}{2^{i+1}}$ patches which are fed to a linear projection layer and a normalization layer for patch embedding. The obtained patch feature maps are flattened into vectors and added with the position embedding before they are  passed through a transformer encoder. The output is reshaped to one feature map. 
The transformer encoder is composed of a spatial-reduction attention layer to reduce the spatial scale of keys and values before the multi-head attention operation and a feed-forward layer \cite{wang2021pyramid}.

\subsubsection{Contextual Attention-based Feature Fusion}\label{sec:ContextualAttentionFeatureFusion}
We design CAFF to utilize the robust multi-scale features collaboratively in the decoder in a pyramid pattern: as illustrated in Fig.\ref{fig:modules}c, a coarser-resolution feature map from the previous scale of the decoder and a finer-resolution feature map from the earlier phase of the encoder are fed to a CAFF module; while the output of this CAFF module and the next finer-resolution feature map from the encoder will be fed to the next CAFF module until the final feature maps are produced (\ie $\frac{W}{4}\times\frac{H}{4}$). In a word, the generated features are incrementally refined in the decoder, and this leads to stronger and more effective tree density estimation.

In each CAFF module, as illustrated in Fig. \ref{fig:modules}c, first, a bilinear interpolation layer is used to upsample a coarser-resolution feature map from the previous scale of the decoder ($S_{i+1}$). A series of convolutional, batch normalization, and ReLU layers are applied to extract tree relevant information from both inputs ($S_{i+1}$, $S_i$). A channel attention (CA) block is devised on the finer-resolution branch ($S_i$) which consists of an average pooling and two fully-connected (FC) layers with a ReLU between them; a sigmoid function is added by the end. Inspired by \cite{hu2018squeeze}, the CA block computes a channel-wise importance vector which is used to multiply with the feature map, so that the tree relevant channels in the feature map are highlighted. 
The re-weighted feature map is finally added with the coarser-resolution feature map ($S_{i+1}$) from the encoder to generate a robust feature map for tree density estimation.

\subsubsection{Tree Density Regressor}\label{sec:TreeDensityRegressor}
The purpose of the TDR module is to estimate the tree density map. The TDR module has been used in three different scales of the decoder to generate tree density maps ($D_1$, $D_2$, and $D_3$ in Fig. \ref{fig:modules}a). The specified scale factor for upsampling the feature maps in the TDR (see Fig.~\ref{fig:modules}d) is set to 1, 2, and 4 respectively for generating the same size of feature maps over scales in the decoder. Afterward, the block of convolutional, batch normalization, and ReLU layers is applied to reduce the number of feature channels and achieve the final density map in each scale. We let every block be responsible for reducing half of the channels (for 128 channels, it is reduced to 1). The original number of feature channels in the first, second, and third decoding scales is 128, 256, and 512, respectively. Hence, we set the number of blocks  ($\tau$ in Fig. \ref{fig:modules}d) in the first, second, and third scales to 1, 2, and 3 correspondingly.

The TDR module is also responsible for perturbing the multi-scale feature maps so that local tree density consistency loss, specified later, will be applied to enforce consistency over multiple density predictions. It applies a perturbation layer before the upsampling layer in the TDR. 
Given a feature map $F$, we specifically choose three types of perturbations including feature perturbation, masking, and spatial dropout from \cite{ouali2020semi} corresponding to $D_1, D_2$ and $D_3$ in Fig.~\ref{fig:modules}a.

\begin{itemize}
\item Feature  perturbation: a noise tensor $\xi \sim U(-0.3, 0.3)$ of the same size as $F$ is uniformly sampled. The noise is injected into $F$ after adjusting the noise amplitude by element-wisely multiplying the noise with $F$, \ie $\tilde {F} = (F\odot \xi) + F$.
\item Feature masking: The sum of $F$ over channels is computed and normalized as $F'$. A mask ($M_{drop}$) is generated by determining a threshold ($\varepsilon \sim U(0.7, 0.9)$) and applying it to $F^{'}$, \ie $M_{drop}=F' \le \varepsilon$. The masked feature map is computed by multiplying $M_{drop}$ to $F$, \ie $\tilde{F} =  F \odot M_{drop}$. In this way, between $10$\%to $30$\% of the most active regions in the feature map are masked. 
\item Spatial dropout: The dropout is applied across the channels of $F$. In other words, some channels are set to zero (dropped-out) and others are activated \cite{tompson2015efficient}. 
\end{itemize}

\subsubsection{Tree Counter Token}\label{sec:TreeCountingToken}
The purpose of the TCT module is to compute the number of trees from Phase 2 to 4 of the encoder (Fig. \ref{fig:modules}a and b). 
In the $i$-th phase, the result of the patch embedding is reshaped to a stack of vectors, $\mathbf{f} = [f_1, f_2, ..., f_{\rho}]$,  $\rho=2^{2(i+1)}$, where each $f$ is a $1 \times C_i$ dimensional feature vector corresponding to a local region. We introduce an additional tree counter token ($f_T$) appended to $\mathbf{f}$, \ie $\mathbf f = [f_1, f_2, ..., f_{\rho}, f_T]$. These vectors are added with positional embedding and  passed through a spatial reduction and multi-head attention blocks in the transformer encoder.
Through the encoding process, $f_T$ aggregates the tree density information from the rest feature vectors in $\mathbf f$ before it is fed to the TCT module to calculate the total number of trees.

In the TCT module, as illustrated in Fig. \ref{fig:modules}e, the tree count is estimated after applying the aforementioned perturbation layer and a convolutional layer.

Since here the input of the perturbation layer is a vector instead of a matrix, the feature masking is performed similarly to the spatial dropout. The difference is that the spatial dropout randomly sets some channels to be zero, while the feature masking selects some of the most active channels to be zero according to the $\varepsilon$.

\subsection{Pyramid Learning strategy}\label{sec:PyramidLearningStrategy}
We design a pyramid learning strategy that consists of three levels such as pixel-level, region-level, and image-level learning to train the TreeFormer. Analyzing the results obtained at different levels of details can increase the accuracy in a coarse-to-fine manner. At the pixel level, the distribution matching loss is used as a supervised loss to evaluate the results for labeled data. At the region level, two losses including local tree density consistency and local tree count ranking are proposed for unlabeled data. 
At the image level, the total number of trees is estimated by the TCT for learning both labeled and unlabeled data.
To clarify, the pyramid learning is not a multi-stage learning but is an end-to-end learning. Pyramid means that the loss functions are defined on different levels of the input while all loss functions are optimized simultaneously.

\subsubsection{Pixel-level learning}\label{sec:PixelLevelLearning}
To optimize the crowd density at the pixel level, the distribution matching loss is utilized \cite{wang2020distribution}. This loss function is based on the combination of the counting loss, optimal transport loss, and total variation loss. The counting loss ($L_c$) calculates the difference between the estimated and ground truth tree density value at the pixel level:
\begin{equation}
    \ L_c = \sum\limits_{k=1}^{K}|\Vert D_k\Vert - \Vert D_{gt} \Vert| 
\end{equation}
where {$K$ is the number of scales in the decoder, $K =3$; $D_k$ is the estimated density map at a certain scale and $D_{gt}$ is the corresponding ground truth.  $\Vert . \Vert$ denotes the $L1$ norm to accumulate the density values in $D_k$ or $D_{gt}$.
The optimal transport loss ($L_{ot}$) calculates the difference between the distribution of the normalized density function of the estimated density map and ground truth \cite{wang2020distribution} as follows:
\begin{equation}
    \ L_{ot}= \sum\limits_{k=1}^{K} W(\frac{D_k}{\Vert D_k \Vert},\frac{D_{gt}}{\Vert D_{gt} \Vert})
\end{equation}
where $W$ is the optimal transport cost referred to~\cite{wang2020distribution}. 
Finally, the total variation loss  $L_{tv}$ is used to stabilize the training procedure, defined as below:  
\begin{equation}
    \ L_{tv}=\sum\limits_{k=1}^{K} \frac{1}{2} \Vert \frac{D_k}{\Vert D_k \Vert}-\frac{D_{gt}}{\Vert D_{gt} \Vert}\Vert
\end{equation}
It is used to alleviate the $L_{ot}$'s poor approximation in the low-density areas.
Accordingly, the overall distribution matching loss for pixel-level learning is formulated as:
 
\begin{equation}
    \ L_{dm}= \alpha _1 L_{c}+\alpha _2 L_{ot}+\alpha _3 L_{tv}
\end{equation}
where $\alpha _i$ is the weight value and is set to 1, 0.1, and 0.01 for $\alpha _1$, $\alpha _2$, and $\alpha _3$, respectively \cite{wang2020distribution}. 

\subsubsection{Region-level learning}\label{sec:RegionLevelLearning}
Our proposed loss function for region-level learning has two parts, \ie local tree count ranking loss and local tree density consistency loss. To implement them, the super- and sub-regions are cropped from the estimated density maps. The cropped regions have the same center and aspect ratio as the original one. 
They are cropped by reducing their size iteratively by a scale factor of 0.75. Below we introduce our loss function upon these regions.

\textbf{Local tree count ranking.} \label{LocalTreeCountRanking}
This learning strategy serves as a self-supervised function that is used for unlabeled images. (Fig. \ref{fig:overview}b). Inspired by \cite{liu2018leveraging}, the number of trees in a super-region is bigger than or at least equal to that of trees in its sub-regions. The network learns the ordinal relation of the cropped density maps by applying a ranking loss: 

\begin{equation}
    \ \gamma = max(0, \vartheta(d_m)-\vartheta(d_n)) 
\label{eq:rank}
\end{equation}
where $d_n$ and $d_m$ are the cropped super- and sub-regions from the estimated density map of an unlabeled image, respectively. $\vartheta$ sums the density values in a region, which signifies the number of estimated trees in this region.  
According to Eq. \ref{eq:rank}, $\gamma$ will be zero when the ordinal relation is correct. We propose a multi-scale structure so that the ranking loss is adopted in the estimated density map of each scale of the decoder (Fig. \ref{fig:overview}).
The loss for each unlabeled image is computed by:
\begin{equation}
    \ L_{rank}=\sum\limits_{k=1}^K \sum\limits_{m=1}^{M-1} \sum\limits_{n=m+1}^M max(0, \vartheta(d_{m,k})-\vartheta(d_{n,k}))
\end{equation}
where $M$ is the number of cropped patches from a density map and $K$ is the number of scale in the decoder. 

\textbf{Local tree density consistency.} The purpose of this strategy is to minimize the discrepancy between predictions at different scales after applying a perturbation to each scale (Fig. \ref{fig:overview}c).  
Since we do not have the ground truth, we use the mean prediction over different scales of the decoder as the pseudo ground truth.
We compute the Kullback–Leibler (KL) divergence between the mean prediction and the prediction at each scale to enforce the network to minimize this distance: %\hamed{I checked some papers, they only used 

\begin{equation}
    \ L_{consis}= \sum\limits_{k=1}^{K} \sum\limits_{m=1}^{M} \sum\limits_{i=1}^{w} \sum\limits_{j=1}^{h} {d_{m,k}}(i,j) \cdot log \frac{{d_{m,k}}(i,j)}{{d_{ave}(i,j)}}
\end{equation}

where $d_{m,k}$ is a certain cropped region from the density map of the $k^{th}$ scale while $d_{avg}=\frac{1}{K}\sum\limits_{k=1}^{K} d_{m,k}$; $i$ and $j$ represent the position of a pixel in the cropped density map with a size of $w\times h$, respectively. Notice we use the same set of cropped regions as for the local tree count ranking. Yet, consistency is applied between the same density regions over different decoding scales.

\subsubsection{Image-level learning}\label{sec:ImageLevelLearning}
The predicted total numbers of  trees from TCT modules at different encoder {phases} are utilized to optimize the network parameters using both labeled and unlabeled data (Fig. \ref{fig:overview}d).

\textbf{Global tree count regularization.} For labeled data, the estimated values by TCTs over three {phases}, $\{t_1^l, t_2^l, t_3^l\}$, are compared with the total number of trees,  $t^l_{gt}$, in the ground truth. For unlabeled data, 
since the ground truth is unavailable, the average of the estimated count values, $\{t_1^{u}, t_2^{u}, t_3^{u}\}$,  $t_{avg}^{u}=\frac{1}{K}\sum\limits_{k=1}^{K} t_k^{u}$, is used as pseudo ground truth to supervise the training. 
The image-level loss functions for labeled ($L_{ts}$) and unlabeled images ($L_{tu}$) are therefore defined by: 

\begin{equation}
    \ L_{ts}=\sum\limits_{k=1}^{K} \Vert t_k^l - t_{gt}^l\Vert
\end{equation}
\begin{equation}
    \ L_{tu}=\sum\limits_{k=1}^{K} \Vert t_k^{u} - t_{avg}^{u}\Vert
\end{equation}

\subsubsection{Training loss}\label{sec:TrainingLoss}
Overall, the loss function for the labeled images is based on the summation of the $L_{dm}$ and $L_{ts}$ ($L_s=L_{dm}+L_{ts}$). The loss function for the unlabeled images comprises three components including $L_{consis}$, $L_{rank}$ and $L_{tu}$ ($L_u=L_{consis}+L_{rank}+L_{tu}$). The final loss is the combination of $L_s$ and $L_u$ with a hyperparameter $\lambda$.
\begin{equation}
    L_=L_{s}+\lambda L_{u}
\end{equation}

\section{Experiments}\label{sec:Experiments}
\subsection{Datasets}\label{sec:Datasets}
\subsubsection{KCL-London dataset}\label{sec:LondonDataset}
This dataset, as specified in Sec.~\ref{DataSource}, contains high-resolution images with 0.2m GSD from London that is divided into two parts including 613 labeled and 308 unlabeled images (Fig. \ref{fig:dataset}). Within the labeled set, we separate it into 452 samples for training and  161 samples for testing. The unlabeled set can be optionally used.

\subsubsection{Jiangsu dataset}\label{sec:JiangsuDataset}
This study area contains 24 Gaofen-II satellite images with 0.8m GSD which are captured from Jiangsu Province, China for training and testing \cite{yao2021tree, liu2021deep}. There are 664,487 trees that are manually annotated across 2400 images. The images cover different landscapes such as cropland, urban residential area, and hill. This dataset is divided into a training set that contains a total of 1920 images, and a test set that contains 480 images. 

\subsubsection{Yosemite dataset}\label{sec:YosemiteDataset}
 This study area is centered at Yosemite National Park, California, United States of America \cite{chen2022transformer}. A rectangular image with 19,200 × 38,400 pixels and 0.12m GSD that is collected from Google Maps and consists of 98,949 trees which are manually annotated. This data is divided into training (1350 images) and test data (1350 images).

{The characteristics of the study areas for different datasets are presented in Table. \ref{tab:DatasetCharacteristics}}.

\begin{table*}[!t]
\centering
	\caption{The characteristic of the utilized datasets.}
	\begin{center}
	\begin{tabular}{c|c c c c c c c c c}
   \toprule
		Dataset name &  Landscape type & Image size & \thead{GSD \\  (m)} & \thead{Number of \\ images}  & \thead{Minimum \\number of trees} & \thead{Maximum \\number of \\ trees} & \thead{Average density \\(tree/image)} & Total\\[5pt]
	\midrule
	    KCL-London &  urban residential area, dense park & $1024\times 1024$ & 0.20 & 613 & 4  & 332 & 155 & 95,067 \\[3pt]
	    Jiangsu &  cropland, urban, residential area & $256\times 256$ & 0.80 & 2400 & 0 & 3132  &  276 & 664,487\\[3pt]
	    Yosemite &  wooded mountainous & $512\times 512$ & 0.12 & 2700 & 0  & 113 &  36 & 98,949 \\[3pt]
	\bottomrule
	\end{tabular}
	\end{center}
    \label{tab:DatasetCharacteristics}	
\end{table*}

\subsection{Evaluation Protocol and Metrics }\label{sec:EvaluationMetrics}
To set up for the semi-supervised experiments, we divide the training set of each data set into 10\% \vs 90\% and 30\% \vs 70\%  for labeled and unlabeled subsets, respectively. We refer to the two settings as \emph{default setting} $1$ and $2$. Notice in the KCL-London dataset, there are also 308 additional unlabeled images (no annotations at all), they can also be used if specified.
For the sake of convenience, we give the notations for different sets in each dataset: first, we denote by $\mathcal D_{tr}$ and $\mathcal D_{te}$ the training and test set, respectively;  and $\mathcal D_{ltr}$ and $\mathcal D_{utr}$ the labeled and unlabeled subset within $\mathcal D_{tr}$; finally, $\mathcal D_{au}$ the additional unlabeled set in the KCL-London dataset.

Following \cite{yao2021tree,chen2022transformer} we use three criteria including mean absolute error ($E_{MAE}$), root mean squared error ($E_{RMS}$), and R-Squared ($E_{R2}$) to evaluate results. 
They are defined as follows: 

\begin{equation}
    \ E_{MAE}=\frac{1}{N}\sum \limits _{i=0}^N |y_i^e-y_i^{gt}|
\end{equation}

\begin{equation}
    \ E_{RMS}=\sqrt{\frac{1}{N}\sum \limits _{i=0}^N (y_i^e-y_i^{gt})^2}
\end{equation}

\begin{equation}
    \ E_{R^2}=1-\frac{\sum \limits _{i=0}^N (y_i^e-y_i^{gt})^2}{\sum \limits _{i=0}^N (y_i^e-\bar{y}^{gt})^2}
\end{equation}
where $N$ denotes the number of samples, 
%$i$ denotes the subscript for a different sample, 
$y_i^e$ represents the estimated tree number for the $i$-th sample, $y_i^{gt}$ is the corresponding ground truth tree number and $\bar{y}^{gt}$ is the mean  ground truth tree number over samples. In general, lower $E_{RMS}$ and $E_{MAE}$ values and higher $E_{R2}$ indicate better performance. 

Besides $E_{RMS}$ and $E_{MAE}$ that only consider the global count at the sample (image) level, we also follow \mbox{\cite{guerrero2015extremely,li2018csrnet}} to employ the grid average mean absolute error (GAME) to analyze the performance of the proposed model at the region-level. GAME typically has four levels including $E_{G0}$, $E_{G1}$, $E_{G2}$, and $E_{G3}$. For a specific level $L$, we subdivide the image into $4^L$ non-overlapping regions, and the estimated tree number is compared with the ground truth tree number in each sub-region:
\begin{equation}
    \ E_{GL}=\frac{1}{N} \sum \limits _{i=0}^N \sum \limits _{l=1}^{4^L} |y_{i,l}^e-y_{i,l}^{gt}|
\end{equation}
where $y_{i,l}^e$ is the estimated tree number in the $l$-th sub-region of the $i$-th image while $y_{i,l}^{gt}$ is the corresponding ground truth. When $L$ increases, the number of subdivided regions increases and the evaluation becomes more subtle.

Moreover, we also follow \mbox{\cite{wang2020nwpu}} to employ the Precision ($E_{P}$), Recall ($E_{R}$), and F1-measure ($E_{F1}$) to assess the performance of the proposed model at the pixel level. They are calculated based on the number of true positives, false positives, and false negatives obtained from the comparison between the predicted density map and the ground truth density map pixel-wisely.

We use $E_{G0}$, $E_{G1}$, $E_{G2}$, $E_{G3}$, $E_{P}$, $E_{R}$, and $E_{F1}$ only for the ablation study to demonstrate the tree localization accuracy by using our proposed region-level and pixel-level optimization strategies.

\subsection{Implementation Details} \label{sec:ImplementationDetails}
We build an encoder-decoder architecture in that the encoder is based on a transformer with four phases. The parameters of the transformer are set according to~\cite{wang2021pyramid}.
The decoder estimates three-scale density maps (Sec.~\ref{sec:TreeFormerFramework}). The number of channels in these scales is 128, 256, and 512 after applying the CAFF modules. The $\tau$ value in TDR (Fig. \ref{fig:modules}d) is set to 1, 2, and 3 for the first, second and third scales of the decoder.

We augment the training set using horizontal flipping and random cropping \cite{amirkolaee2019height}. Also, we randomly crop the images with a fixed size of 256×256 as the input of the network. The number of the epoch, batch size, learning rate, and weight decay are set to 500, 16, $10^{-4}$, and $10^{-5}$, respectively. The Adam optimizer is used. All parameters are tuned on the KCL-London dataset and utilized for all experiments. The ground truth contains the coordinates of tree locations which are specified by annotations dots. We follow \cite{zhang2016single} to generate the ground truth density maps from the tree locations using Gaussian functions.

\subsection{Comparisons with state of the art} \label{sec:ComparisonsSOTA}
 In this section, the purpose is to evaluate the performance of the proposed TreeFormer with state of the art models. We categorize comparisons into two groups including semi-supervised ones and supervised ones. 
  
\textit{Semi-supervised models}: Comparable models in the semi-supervised group are all trained under our default setting. To this end, four state of the art semi-supervised methods  including cross-consistency training (CCT) \cite{ouali2020semi}, mean-teacher (MT) \cite{tarvainen2017mean}, interpolation consistency training (ICT) \cite{verma2022interpolation}, and learning to rank (L2R) \cite{liu2018leveraging} are selected. These methods were not originally proposed for tree counting while we adapt them into our task for comparison. For instance, the CCT, ICT, and MT were originally proposed for the image classification task while we adapt them to predict density maps and change their image-level classification consistency loss to our proposed local tree density consistency loss. The L2R was originally proposed for crowd counting and we transfer it to tree counting; L2R only uses a single ranking loss on the final prediction while our local tree count ranking loss is defined over multiple perturbed intermediate scales of the decoder.

For the comparison, we use 10\% or 30\% of training data as $\mathcal D_{ltr}$ while the rest as $\mathcal D_{utr}$. 
To make a fair comparison, the same transformer blocks are used as the backbone for comparable methods. 
Table \ref{tab:semisuperviseT} shows that our TreeFormer significantly outperforms others under the same level of supervision. For instance, on the KCL-London dataset, for 10\% and 30\% labeled data, we observe a decrease of 2.87 and 3.60 for $E_{MAE}$, 3.78 and 4.55 for $E_{MSE}$, and an increase of 0.15 and 0.11 for $E_{R^2}$ from TreeFormer to CCT. On the Jiangsu dataset, our model also has 18.54 and 9.35 decreases of $E_{MAE}$ to the previously best-performed model CCT using 10\% and 30\% labeled data, respectively. The same observation also goes for the Yosemite dataset. 

Overall, our model produces the lowest errors on the Yosemite dataset amongst all datasets. We believe the reason lies in the simple image characteristics obtained in this study area (see Table~\ref{tab:DatasetCharacteristics}).
In the Yosemite dataset, the background and trees are very different, which makes tree identification easier. While in the KCL-London and Jiangsu datasets, there are various objects such as buildings, cars, vegetation, \etc, which makes the identification and counting of trees challenging. Also, in the Jiangsu dataset, the lower resolution of the images compared to KCL-London has reduced the accuracy of its results. In Fig. \ref{fig:SemisuperviseF}, we show some qualitative results of our method compared with other semi-supervised methods.

\begin{table*}[!t]
\centering
	\caption{Comparison with state of the art semi-supervised methods on KCL-London, Jiangsu, and Yosemite datasets. 
 The best and second results are marked in red and blue, respectively.}
	\begin{center}
	\begin{tabular}{c c|c c c|c c c|c c c}
   \toprule
    {}  & Dataset & \multicolumn{3}{c}{KCL-London}  & \multicolumn{3}{c}{Jiangsu} & \multicolumn{3}{c}{Yosemite} \\
	\midrule
		Setting & Method & $E_{MAE} \downarrow$ & $E_{RMS}  \downarrow$ & $E_{R^2} \uparrow$ & $E_{MAE} \downarrow $ & $E_{RMS} \downarrow$  & $E_{R^2} \uparrow$ & $E_{MAE} \downarrow$ & $E_{RMS} \downarrow$ & $E_{R^2} \uparrow$ \\
	\midrule
        {} & MT \cite{tarvainen2017mean} & 34.03 & 43.09 & 0.13 & 99.07 & \textcolor{blue}{158.19} & \textcolor{blue}{0.76} & 10.97 & 13.70 & 0.51\\
	  {10\%Labeled } &  ICT \cite{verma2022interpolation} & 37.83 &  46.80 & 0.09 & 110.34 & 245.56 & 0.54 & 8.70 & 11.40 & 0.66\\  
	    {90\%Unlabeled} & L2R \cite{liu2018leveraging}  & 34.86 & 42.05 & 0.17 &  100.89 & 198.60 & 0.62 & 7.56 & 10.08 & 0.73\\
        {} &  CCT \cite{ouali2020semi} & \textcolor{blue}{30.70} & \textcolor{blue}{39.63} & \textcolor{blue}{0.26} & \textcolor{blue}{91.86} & 158.82 & 0.75 & \textcolor{blue}{6.76} & \textcolor{blue}{8.95} & \textcolor{blue}{0.79}\\
        {} &  TreeFormer & \textcolor{red}{27.83} & \textcolor{red}{35.85} & \textcolor{red}{0.41} & \textcolor{red}{73.32} & \textcolor{red}{119.36} & \textcolor{red}{0.86} & \textcolor{red}{5.72} & \textcolor{red}{7.63} & \textcolor{red}{0.84}\\
	\midrule
	    {} & MT \cite{tarvainen2017mean} & 26.53 & 34.63 & 0.44 & 79.82 & 129.37 & 0.84 & 8.79 & 11.49 & 0.59\\
	  {30\%Labeled}  & ICT \cite{verma2022interpolation} & 32.27 & 39.94 & 0.11 & 88.65 & 167.08 & 0.68 & 6.86 & 9.21 & 0.78\\
        {70\%Unlabeled} & L2R \cite{liu2018leveraging} & 24.71 & 31.96 & 0.52 & 70.72 & 121.40 & 0.86 & \textcolor{blue}{5.79} & \textcolor{blue}{7.69} & \textcolor{blue}{0.84}\\
	  {}  & CCT \cite{ouali2020semi} & \textcolor{blue}{24.21} & \textcolor{blue}{31.34} & \textcolor{blue}{0.55} & \textcolor{blue}{65.73} & \textcolor{blue}{116.67} & \textcolor{blue}{0.87} & 5.96 & 7.78 & \textcolor{blue}{0.84}\\
        {} & TreeFormer & \textcolor{red}{20.61} & \textcolor{red}{26.79} & \textcolor{red}{0.66} & \textcolor{red}{56.38} & \textcolor{red}{96.34} & \textcolor{red}{0.91} & \textcolor{red}{4.69} & \textcolor{red}{6.26} & \textcolor{red}{0.89}\\       
	\bottomrule
	\end{tabular}
	\end{center}
    \label{tab:semisuperviseT}	
\end{table*}

\begin{figure*}[!t]
\centering
\includegraphics[width=7in]{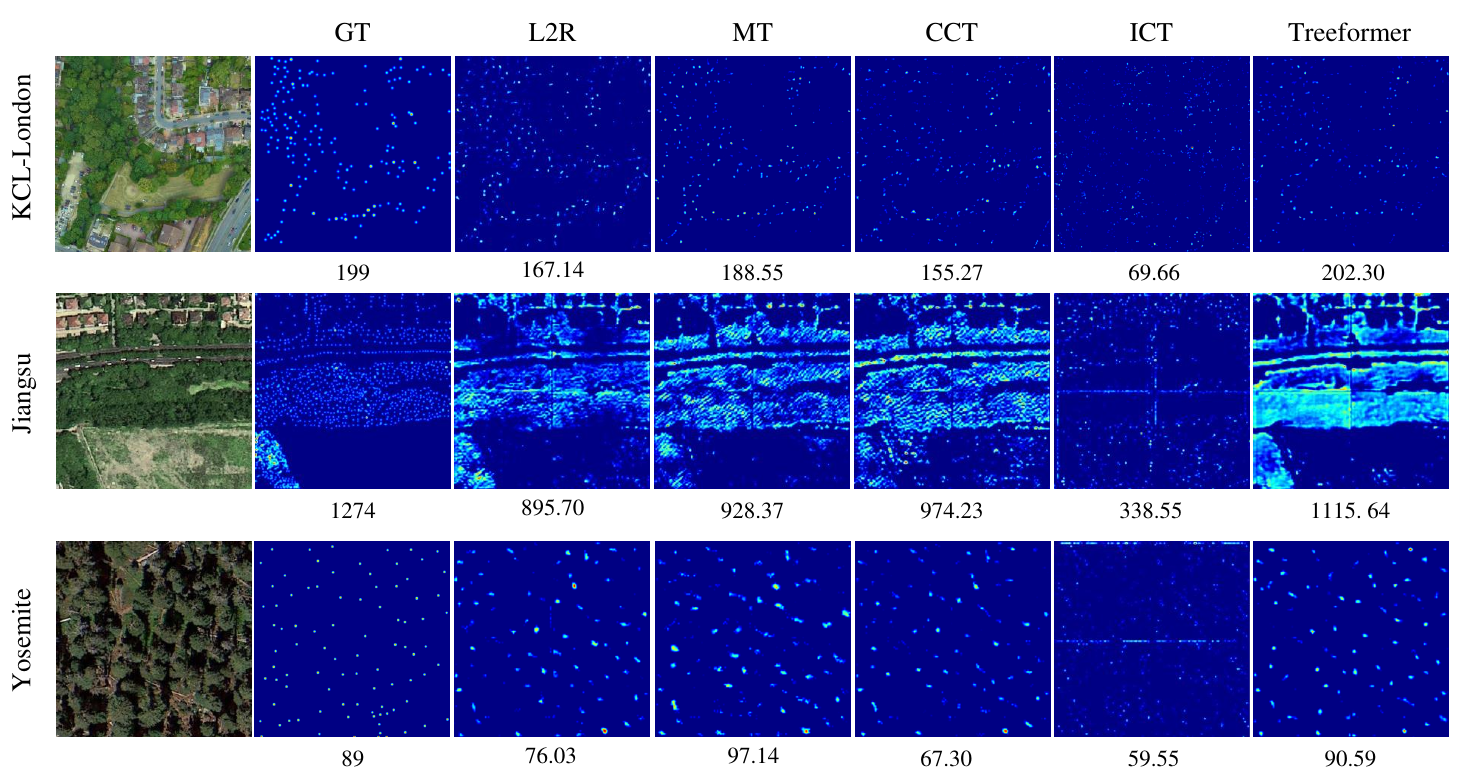}
\caption{Qualitative results of our TreeFormer compared with other semi-supervised methods including CCT \cite{ouali2020semi}, MT \cite{tarvainen2017mean}, ICT \cite{verma2022interpolation}, and L2R \cite{liu2018leveraging} on KCL-London, Jiangsu, and Yosemite datasets. The first column shows sample images of three utilized datasets. The remaining columns show density maps of ground truth (GT) and semi-supervised methods.}

\label{fig:SemisuperviseF}
\end{figure*}

\textit{Supervised models}: To further investigate the effectiveness of our model, we evaluate the proposed model in the case of supervised training and compare it with existing methods (Table \ref{tab:superviseT}). In this scenario, the entire $\mathcal D_{tr}$ is assumed labeled for training. 
%\miaojing{imporatnly here, what would be the number of labeled data ?}
We denote the supervised version of our method by S-TreeFormer, which has the same backbone as the original TreeFormer.  The DM loss and global tree count regularization is still used in the supervised form. The local tree count ranking and local density consistency are however no longer used. The comparable methods include SASNet \cite{song2021choose}, FusionNet \cite{ma2022fusioncount}, EDNet \cite{yao2021tree}, Swin-UNet \cite{cao2021swin}, CSRNet \cite{li2018csrnet}, MCNN \cite{zhang2016single}, DENT \cite{chen2022transformer}, and TreeCountNet \cite{liu2021deep}. Specifically, the SASNet, CSRNet, MCNN, and FusionNet are state of the art crowd counting methods, we reproduce them in the tree counting task. The Swin-UNet is based on the transformer architecture and the others are based on convolutional architecture. The DENT employs a convolutional architecture for extracting the feature maps from the input image. Then a transformer encoder is used to model the interaction of the extracted features and estimate the tree density map.

In the experiment of the KCL-London and Yosemite datasets, our model achieves the highest accuracy. For the Jiangsu dataset, our model obtains the lowest $E_{MAE}$ and $E_{MSE}$, while the TreeCountNet achieves a slightly higher $E_{R^2}$ (0.01). In Fig. \ref{fig:SuperviseF}, we show some qualitative results of our method compared with other supervised methods.

In the last row of Table \ref{tab:superviseT}, the performance of the proposed TreeFormer is presented when the additional unlabeled images in $\mathcal D_{au}$ are used along with all labeled images in $\mathcal D_{tr}$ for network training. Using unlabeled data can further reduce the value of $E_{MAE}$ and $E_{MSE}$ by 1.82 and 1.34, respectively.

\begin{table*}[!t]
\centering
	\caption{Comparison with state of the art supervised methods on KCL-London, Jiangsu, and Yosemite datasets.}
	\begin{center}
	\begin{tabular}{c|c|c c c|c c c|c c c}
   \toprule
    \multicolumn{2}{c}{Dataset}  & \multicolumn{3}{c}{KCL-London}  & \multicolumn{3}{c}{Jiangsu} & \multicolumn{3}{c}{Yosemite} \\[5pt]
	\midrule
		Method & \thead{Unlabeled \\ Images} & $E_{MAE} \downarrow$ & $E_{RMS}  \downarrow$ & $E_{R^2} \uparrow$ & $E_{MAE} \downarrow$  & $E_{RMS} \downarrow$  & $E_{R^2} \uparrow$ & $E_{MAE} \downarrow$ & $E_{RMS} \downarrow$ & $E_{R^2} \uparrow$ \\[5pt]
	\midrule
	    MCNN \cite{zhang2016single} & \xmark &  25.87 & 34.12 & 0.45 & 81.09 & 125.45 & 0.84 & 10.44 & 12.45 & 0.61\\[3pt]
	    CSRNet \cite{li2018csrnet}  & \xmark &  \textcolor{blue}{23.27} & \textcolor{blue}{29.62} & \textcolor{blue}{0.59} & 47.22 & 83.14 & 0.93 & 9.34 & 11.48 & 0.65\\[3pt]
        Swin-UNet \cite{cao2021swin} & \xmark &  36.45 & 47.56 & 0.24 & 70.17 & 110.46 & 0.88 & 13.34 & 16.09 & 0.52\\[3pt]
	    FusionNet \cite{ma2022fusioncount} & \xmark &  28.45 & 35.67 & 0.47 & 54.75 & 89.45 & 0.92 & 6.88 & 9.10 & 0.78\\[3pt]
	    SASNet \cite{song2021choose} & \xmark &  24.33 & 30.12 & 0.56 & 47.32 & \textcolor{blue}{76.90} & 0.94 & \textcolor{blue}{6.33} & \textcolor{blue}{8.46} & \textcolor{blue}{0.81}\\[3pt]
	    EDNet \cite{yao2021tree} & \xmark &  26.18 & 32.02 & 0.52 & 99.10 & 153.47 & 0.77 & 9.92 & 12.39 & 0.60\\[3pt]
		DENT \cite{chen2022transformer} & \xmark & - & - & - & - & - & - & 7.50 & 12.30 & -\\[3pt]
		TreeCountNet \cite{liu2021deep} & \xmark &  - & - & - & \textcolor{blue}{45.08} & 77.96 & \textcolor{red}{0.96} & - & - & -\\[3pt]
		S-TreeFormer  & \xmark &  \textcolor{red}{18.52} & \textcolor{red}{24.32} & \textcolor{red}{0.72} & \textcolor{red}{41.06} & \textcolor{red}{72.06} & \textcolor{blue}{0.95} & \textcolor{red}{4.29} & \textcolor{red}{5.85} & \textcolor{red}{0.91}\\[3pt]
    \midrule
		TreeFormer  & \cmark &   \cellcolor{lightgray}16.70 & \cellcolor{lightgray}22.98 & \cellcolor{lightgray}0.75 & - & - & - & - & - & -\\[3pt]
	\bottomrule
	\end{tabular}
	\end{center}
    \label{tab:superviseT}	
\end{table*}

\begin{figure*}[!t]
\centering
\includegraphics[width=7in]{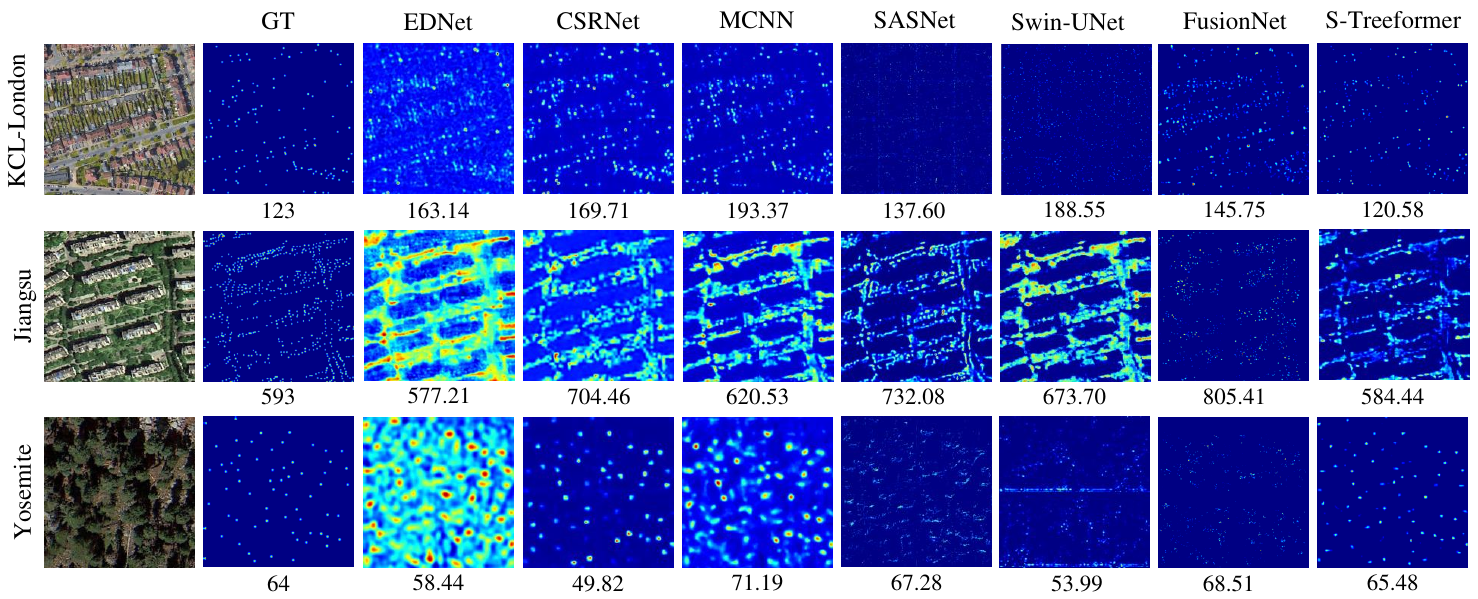}
\caption{Qualitative results of S-TreeFormer compared with other supervised methods including  EDNet \cite{yao2021tree}, CSRNet \cite{li2018csrnet}, MCNN \cite{zhang2016single}, SASNet \cite{song2021choose}, Swin-UNet \cite{cao2021swin}, FusionNet \cite{ma2022fusioncount} on KCL-London, Jiangsu, and Yosemite datasets. The first column shows sample images of three utilized datasets. The remaining columns show density maps of GT and supervised methods.}

\label{fig:SuperviseF}
\end{figure*}

\medskip

Finally, the number of parameters, FLOPS, and inference time of the proposed TreeFormer are compared with other semi-supervised methods in Table \mbox{\ref{tab:NetworkParameters}}. To make a fair comparison, we set the batch size to 1 for all methods on the KCL-London dataset.
According to the results in Table \mbox{\ref{tab:semisuperviseT}}, CCT achieves the second best performance in general, yet it has clearly consumed more FLOPS, parameters, and inference time than the proposed TreeFormer. ICT in general has the lowest computation cost, yet ours compared to ICT is not significantly different. Note that MT and ICT have the same basic architectures, hence their corresponding values in Table \mbox{\ref{tab:NetworkParameters}} are the same. Also, the same observation goes for our TreeFormer and its supervised version S-TreeFormer.

\begin{table}[!t]
\centering
	\caption{Comparison of the FLOPS, number of parameters and inference time between our work and state of the art.}
	\begin{center}
	\begin{tabular}{c|c c c}
   \toprule
		Method &  \thead{FLOPS \\ (G)}  & \thead{Number of  \\ parameters (M)}  & \thead{Inference  \\ time (ms) }\\[5pt]
	\midrule
	    MT &  28.385 & 114.182	 &  13\\[3pt]
	    ICT &  28.385 & 114.182 & 13\\[3pt]
	    L2R  &  34.245 & 115.651 & 15\\[3pt]
	    CCT  &  45.060 & 144.138 & 25 \\[3pt]
	    TreeFormer  &  36.296 & 116.144 & 16 \\[3pt]
	    S-TreeFormer  &  36.296 & 116.144 & 16\\[3pt]
	\bottomrule
	\end{tabular}
	\end{center}
    \label{tab:NetworkParameters}	
\end{table}

\subsection{Ablation Study} \label{sec:AblationStudy}
We analyze TreeFormer on the KCL-London dataset by ablating its proposed components to evaluate their effects on the model accuracy. 
The ablation study is operated on our default semi-supervised setting 2, \ie  30\% labeled images \vs 70\% unlabeled images. 

\subsubsection{Analysis on model architecture} \label{sec:AnalysisModelArchitecture}
In this section, we investigate the proposed PTFR, CAFF and TDR modules. 

\textit{PTFR module.} The pyramid structure of the PTFR can be downgraded by reducing the number of phases of the encoder from 4 into 2 (phases 1 and 2 in Fig. \ref{fig:modules}a) so that only one scale is produced in the decoder. This phase reduction increases $E_{MAE}$ and $E_{RMS}$by 6.78 and 9.80 and reduces the value of $E_{R^2}$ by 0.28 compared to the original TreeFormer.

\textit {CAFF module.} To investigate the effect of the CAFF module, we present the result without using it (w/o CAFF) in Table \ref{tab:CAFF}. It shows that the error value of $E_{MAE}$ is increased by 5.08 when the CAFF module is removed. Next, we devise another variant, using CAFF without channel attention (CAFF w/o CA), which increases the $E_{MAE}$ by 3.92 and $E_{MSE}$ by 7.99. In addition, the effect of changing the channel attention layer to the spatial attention layer (CAFF w/ SA) and using both channel and spatial attention blocks simultaneously (CAFF w/ SA+CA) has also been reported in Table \ref{tab:CAFF}. Accordingly, replacing channel attention with other ways has reduced the accuracy of the results.

\begin{table}[!t]
\centering
	\caption{Ablation study of the CAFF module on KCL-London dataset.}
	\begin{center}
	\begin{tabular}{c|c c c}
   \toprule

		Method &  $E_{MAE} \downarrow$ & $E_{RMS}  \downarrow$ & $E_{R^2} \uparrow$  \\[5pt]
	\midrule
	    w/o CAFF &  27.32 & 34.78 & 0.55\\[3pt]
	    CAFF w/o CA &  24.52 & 30.56 & 0.59\\[3pt]
	    CAFF w/ SA &  23.86 & 29.46 & 0.61\\[3pt]
	    CAFF w/ SA+CA  &  24.12 & 29.34 & 0.60 \\[3pt]
	    TreeFormer  &  \textbf{20.61} & \textbf{26.79} & \textbf{0.66} \\[3pt]

	\bottomrule
	\end{tabular}
	\end{center}
    \label{tab:CAFF}	
\end{table}

\textit {TDR module.} First, the number of blocks of Conv, BN, and ReLU layers in the TDR module ($\tau$ in Fig. \ref{fig:modules}d) is studied for different scales. In Table \ref{tab:TDR}, in the first case, one block of these layers ($\tau$=1) is utilized for all three scales to reduce the channel numbers and calculate the density maps. In the second and third cases, $\tau$ was set to 2 and 3 for all scales, respectively. In the fourth case, $\tau$ was set to 1, 2, and 3 for the first, second, and third scales, respectively (see Fig. \ref{fig:modules}a). The result show that the fourth case achieves the best results. According to Sec.~\ref{sec:TreeDensityRegressor}, the fourth case is also the best choice theoretically.  

Moreover, we analyze the selection of perturbations including feature perturbation ($P_1$), feature masking ($P_2$), Dropout ($P_3$), and  for estimating the tree density maps. 
By default we use $P_1$, $P_2$, and $P_3$ for $D_1$, $D_2$, and $D_3$ (Fig.\ref{fig:overview}a), respectively. Applying the mentioned perturbations has different effects on different scales due to the type of change they produce on the feature maps. 
For instance, applying $P_1$ to $D_3$ would result into more noise than to $D_2$ and $D_1$, because the resolution of $D_3$ (before upsampling) is smaller than that of $D_2$ and $D_1$. Altering too much information in a scale causes network performance drop. Hence, we specifically design $P_1$, $P_2$, and $P_3$ to suit the scale $D_1$, $D_2$, and $D_3$ from fine to coarse. 
We compare it to random order or other specific orders of perturbations in Table~\ref{tab:Perturbation}. The results show that the order $P_1$, $P_2$, $P_3$ works the best.

\begin{table}[!t]
\centering
	\caption{Ablation study of the hyperparameter $\tau$ in the TDR module on KCL-London dataset.}
	\begin{center}
	\begin{tabular}{c|c c c}
   \toprule

		Method &  $E_{MAE} \downarrow$ & $E_{RMS}  \downarrow$ & $E_{R^2} \uparrow$  \\[5pt]
	\midrule
	    TDR ($\tau=1$)  &  26.32 & 32.65 & 0.51 \\[3pt]
	    TDR ($\tau=2$) &  23.76 & 30.42 & 0.55\\[3pt]
	    TDR ($\tau=3$) &  22.45 & 29.74 & 0.58\\[3pt]
	    TDR ($\tau=1,2,3$) &  \textbf{20.61} & \textbf{26.79} & \textbf{0.66}\\[3pt]

	\bottomrule
	\end{tabular}
	\end{center}
    \label{tab:TDR}	
\end{table}

\begin{table}[!t]
\centering
	\caption{Ablation study of the perturbation in the TDR module on KCL-London dataset. The order of the perturbations corresponds to that of $D_1, D_2$ and $D_3$ in Fig.~\ref{fig:modules}a. }
	\begin{center}
	\begin{tabular}{c|c c c}
   \toprule

		Method &  $E_{MAE} \downarrow$ & $E_{RMS}  \downarrow$ & $E_{R^2} \uparrow$  \\[5pt]
	\midrule
	    
	    Random & 25.73 & 34.69 & 0.43 \\[3pt]
        $P_3, P_2, P_1$ & 28.48 & 34.72 & 0.44 \\[3pt]
	    $P_3, P_1, P_2$ & 31.78 & 39.84 & 0.37\\[3pt]
	    $P_2, P_3, P_1$ & 25.43 & 33.68 & 0.49\\[3pt]
	    $P_2, P_1, P_3$ & 24.92 & 32.81 & 0.53\\[3pt]
	    $P_1, P_3, P_2$ & 22.25 & 29.65 & 0.60\\[3pt]
	    $P_1, P_2, P_3$ & \textbf{20.61} & \textbf{26.79} & \textbf{0.66}\\[3pt]
	\bottomrule
	\end{tabular}
	\end{center}
    \label{tab:Perturbation}	
\end{table}

\subsubsection{Analysis on learning strategy}\label{sec:AnalysisLearningStrategy}
We introduce a pyramid learning strategy that consists of three levels such as pixel-level, region-level, and image-level learning. 

\textit{Pixel-level learning.} To verify that the designed strategy is effective, we utilize the $L2$ loss instead of the DM loss (w/ L2). Table \ref{tab:LearningStrategy} shows that using $L2$ increases the $E_{MAE}$ by 18.62 and $E_{RMS}$ by 25.64. Also, the $E_{P}$, $E_{R}$, and $E_{F1}$ as the pixel-level localization metrics exhibit a reduction of 27.86\%, 45.75\%, and 38.55\%, respectively.

\begin{table*}[t]
\centering
	\caption{Ablation study of the learning strategy on KCL-London dataset.}
	\begin{center}
	\begin{tabular}{c|c c c c c c c c c}
   \toprule
		Method &  $E_{MAE} \downarrow$ & $E_{MSE} \downarrow$ & $E_{R^2} \uparrow$ &  $E_{G1}  \downarrow$ & $E_{G2} \downarrow$ & $E_{G3} \downarrow$ & $E_{P} (\%) \uparrow$ & $E_{R}  (\%)\uparrow$ & $E_{F1}  (\%) \uparrow$ \\[5pt]
	\midrule
	    w/ $L2$ &  39.23 & 52.43 & 0.26 & 47.65 & 63.12 & 89.71 & 24.15 & 14.66 & 17.35\\[3pt]
	    w/ LTC-JS &  22.56 & 28.80 & 0.63 & 31.05 & 46.64 & 71.12 & 50.83 & 57.68 & 54.22\\[3pt]
	    w/o LTC &  24.91 & 34.12 & 0.59 & 34.12 & 49.81 & 73.43 & 47.96 & 53.16 & 50.44\\[3pt]
	    w/ STC &  22.46 & 32.55 & 0.61 & 32.55 & 48.01	& 71.91	& 50.15 & 55.67 & 52.63\\[3pt]
	    w/o LTR  &  26.24 & 36.21 & 0.57 & 36.21 & 51.34 & 74.21 & 46.56 & 49.23 & 47.32 \\[3pt]
	    w/ STR  &  24.63 & 34.23 & 0.58 & 34.23 & 49.24 & 73.24 & 48.31 & 53.86 & 50.76 \\[3pt]
	    w/o GTC  &  21.78 & 28.92 & 0.62 & 29.91 & 46.11 & 70.86 & 51.45 & 59.94 & 55.12\\[3pt]
	    TreeFormer  &  \textbf{20.61} & \textbf{26.79} & \textbf{0.66} & \textbf{29.04} & \textbf{45.18} & \textbf{70.49} & \textbf{52.01} & \textbf{60.41}	& \textbf{55.90}\\[3pt]
	\bottomrule
	\end{tabular}
	\end{center}
   \label{tab:LearningStrategy}	
\end{table*}

\textit{Region-level learning}. Investigating the performance of the proposed TreeFormer without the local tree density consistency (w/o LTC) indicates an increase of $E_{MAE}$ by 4.30 and $E_{RMS}$ by 6.50 {(Table \ref{tab:LearningStrategy})}. Furthermore, analyzing the computed region-level metrics shows that the $E_{G1}$, $E_{G2}$, and $E_{G3}$ are increased by 5.08, 4.63, and 2.84, respectively.
The consistency is applied over different cropped regions of the image. If we only apply it on the single  image level (w/ STC), the performance will also be improved compared to that w/o LTC. However, applying consistency on cropped regions clearly leads to better accuracy.

LTC employs the KL divergence to measure the distance between the obtained density distribution from unlabeled images and that from the ground truth. A variant is to use the Jensen-Shannon (JS) divergence, which measures the total distance from any one distribution to the average of the two probability distributions. We compare the results of using KL divergence and JS divergence in Table \mbox{\ref{tab:LearningStrategy}} (TreeFormer \vs w/ LTC-JS), which shows the better performance of using KL divergence in estimating tree densities. KL is more suitable than JS in our task: KL divergence is asymmetric, which means it measures the difference between two density maps in one direction only \mbox{\cite{wu2022robust,wang2017multi}}. This makes it suitable in density estimation task where one density map is known to be a reference. In contrast, JS divergence is symmetric, it treats both density maps as equal. 

On the other hand, the performance of the model without using the proposed local tree count ranking loss is also evaluated (w/o LTR). {Table \ref{tab:LearningStrategy} shows that} the $E_{MAE}$ and $E_{MSE}$ are increased by 5.63 and 10.64, respectively. Besides, not using the LTR also reduces the region-level accuracy and increases the $E_{G1}$, $E_{G2}$, and $E_{G3}$ by 7.17, 6.16, and 3.72, respectively. We also present a variant by utilizing a single ranking loss only for the last layer ($D_1$ in Fig. \ref{fig:overview}a) of the model (w/ STR). This has a weaker performance than applying it on the intermediate scales of the decoder.

\textit{Image-level learning.}
At last, to show the advantage of using the proposed global tree count regularization, the performance of TreeFormer without using it (w/o GTR) is evaluated. Table \ref{tab:LearningStrategy} demonstrates that the $E_{MAE}$ and $E_{MSE}$ are increased by 1.17 and 2.13, respectively.

\subsubsection{Analysis of the effect of the number of labeled images}\label{sec:AnalysisNumberLabel}
In this section, TreeFormer is examined in the case of using different amounts of labeled training data. In the first evaluation, 10\% of labeled and 90\% of unlabeled images are used for network training. 
This process is carried on for 20\%, 30\%, 40\%, 60\%, 80\%, and 100\% of labeled training data with the rest percent being the unlabeled training data, and the obtained values are shown in Fig. \ref{fig:plot}a. It can be seen that the counting accuracy using TreeFormer with 30\% labeled images is already close to the fully-supervised model (the star point). Moreover, the performance of the network in supervised form without using the unlabeled images (S-TreeFormer, see Sec.~\ref{sec:ComparisonsSOTA}) is also assessed (Fig. \ref{fig:plot}a). One can also see a big error reduction between S-TreeFormer and TreeFormer, which verifies the effectiveness of our semi-supervised framework.

Next, we further investigate Treeformer by fixing 10\% of labeled images while gradually adding more unlabeled images. In Fig. \ref{fig:plot}b we present the result with unlabeled images increased from 100 to 700 images (including the images in $\mathcal D_{au}$). The $E_{MAE}$ keeps decreasing in this process. 

Afterward, the number of the fixed labeled images is increased to 50\%, while the number of unlabeled images is gradually added from 100 to 500 (Fig. \mbox{\ref{fig:plot}c}). It can be seen that the $E_{MAE}$ is decreased from 21.4 using 100 unlabeled images to 19.4 using 500 unlabeled images.

Finally, all labeled images are used and the number of unlabeled images is increased from 100 to 308. According to Fig.\mbox{\ref{fig:plot}d}, the $E_{MAE}$ is decreased from 17.6 using 100 unlabeled images to 16.6 using 308 unlabeled images.

Overall, by choosing a small number of labeled training data as opposed to a large number of unlabeled data, the effect of using more unlabeled data on the accuracy of the results becomes more apparent.

\begin{figure}[!t]
\centering
\includegraphics[width=3.5 in]{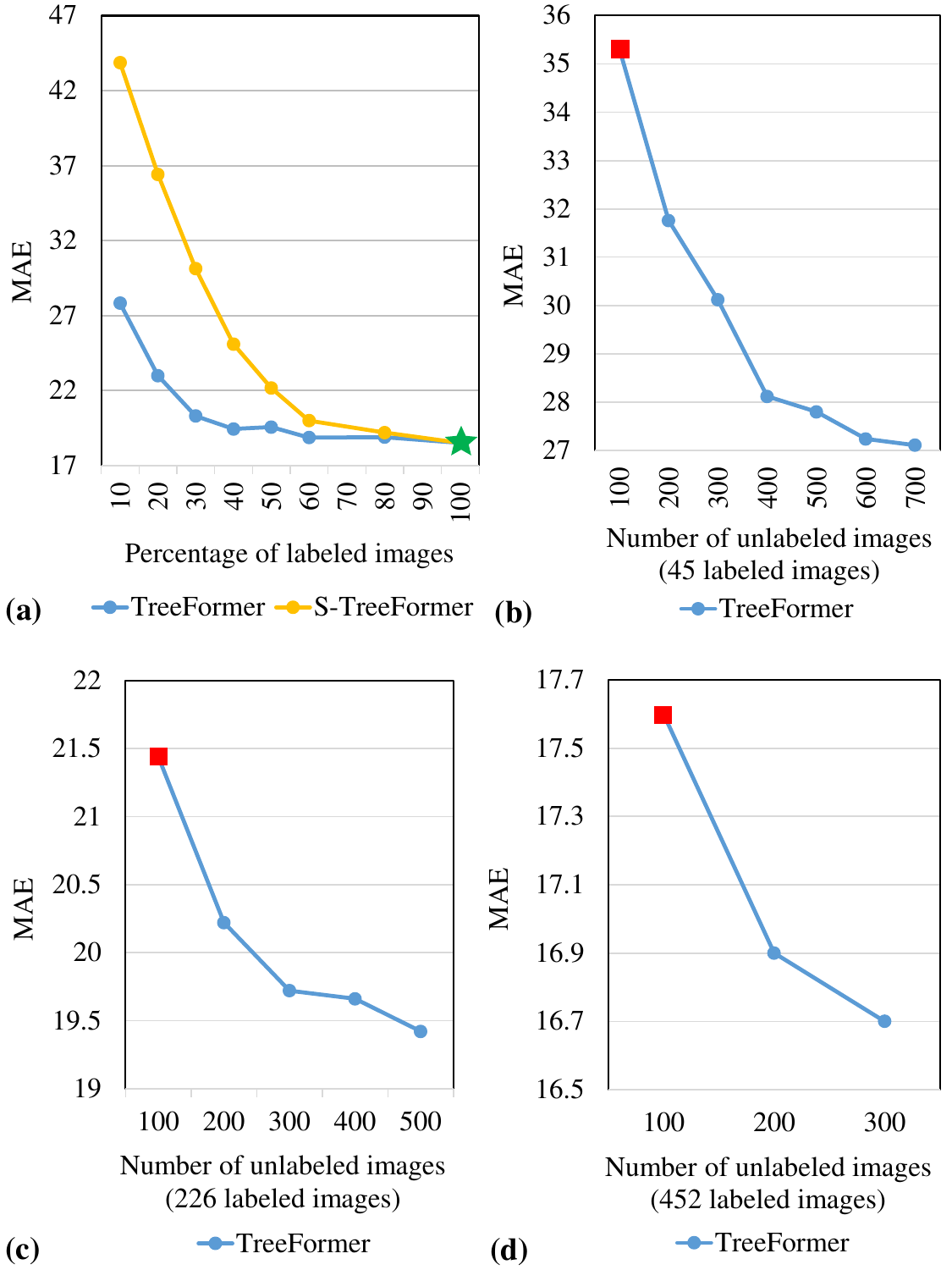}
\caption{(a) The trend of $E_{MAE}$ changes in the S-TreeFormer and TreeFormer with the increase of the percentage of labeled images. The trend of $E_{MAE}$ changes in the TreeFormer with the increase of the number of unlabeled images when fixing (b) 10\%, (c) \%50, and (d) \%100 of labeled images.}
\label{fig:plot}
\end{figure}

\section{Conclusion}\label{sec:Conclusion}
In this paper, we propose a semi-supervised architecture based on transformer blocks for tree counting from single remote sensing images. In this network, the contextual attention-based feature fusion module is introduced to combine the extracted features during the encoding process with the decoding part of the network. In addition, the tree density regressor module is designed to estimate the tree density map after applying different perturbations. The tree counter token is introduced to calculate the total number of trees in the encoding phases and the obtained global count plays the role of the regulator to improve training performance. Moreover, we propose a  pyramid learning strategy that includes local tree count ranking and local tree density consistency to leverage unlabeled images into the training. 

A new tailored tree counting dataset, KCL-London, is constructed with Google Earth images and the locations of the central points of tree canopies were annotated manually. The results on three datasets demonstrate that our method achieves superior performance compared with the state of the art in semi-supervised and supervised tasks.

Counting trees has multiple applications in environmental intelligence and environmental management. The algorithm developed here is scalable to a range of commonly available high-resolution image types. 
Accessibility of open source high-resolution imagery is fundamental to being able to map and therefore manage trees in both urban and rural areas. 

Trees in different regions of the earth have different and varied shapes and canopies. It is not realistic to prepare training data from all available domains for network training, hence, improving the generalizability of the proposed network on the heterogeneous dataset (e.g. NEON dataset \mbox{\cite{weinstein2020neon}}) by domain generalization and adaptation techniques can be the future work.
\section*{Acknowledgements}
This project (ReSET) has received funding from the European Union’s Horizon 2020 FET Proactive Programme under grant agreement No 101017857. The contents of this publication are the sole responsibility of the ReSET consortium and do not necessarily reflect the opinion of the European Union. Miaojing Shi was also supported by the Fundamental Research Funds for the Central Universities.

\bibliographystyle{IEEEtran}
\bibliography{Citation}

\end{document}